\documentclass[runningheads]{llncs}
\usepackage{graphicx}
\begin{document}
\title{Federated Learning Meets Natural Language Processing: A Survey}
%
%
%
\author{Ming Liu, Stella Ho, Mengqi Wang, Longxiang Gao, Yuan Jin, He Zhang}
\authorrunning{L. Ming et al.}
%
\institute{Deakin University}
%
\maketitle              
\begin{abstract}
Federated Learning aims to learn machine learning models from multiple decentralized edge devices (e.g. mobiles) or servers without sacrificing local data privacy. Recent Natural Language Processing techniques rely on deep learning and large pre-trained language models. However, both big deep neural and language models are trained with huge amounts of data which often lies on the server side. Since text data is widely originated from end users, in this work, we look into recent NLP models and techniques which use federated learning as the learning framework. Our survey discusses major challenges in federated natural language processing, including the algorithm challenges, system challenges as well as the privacy issues. We also provide a critical review of the existing Federated NLP evaluation methods and tools. Finally, we highlight the current research gaps and future directions.

\end{abstract}
\section{Introduction}
Modern machine learning algorithms rely on big amounts of data, especially when training deep neural models from high dimensional data such as text and image. Most data naturally come from end users, which are distributed and separated by different end devices. It is necessary to learn well performed machine learning models while preserving users' privacy. Federated learning (FL) has become a new machine learning paradigm to train a model across multiple decentralized edge devices or servers holding local data samples without exchanging them. The term \textit{federated learning} was first proposed in 2016 by \cite{DBLP:journals/corr/McMahanMRA16}: "We term our approach Federated Learning, since the learning task is solved by a loose federation of participating devices (which we refer to as clients) which are coordinated by a central server." In the real world scenario, organizations such as different hospitals hold confidential data, while these hospitals would like to train a disease classification model for common use, it is hard to ask them to upload their own data to the cloud. Even within the same hospital, different departments often save patients' information locally. Another example is human beings create lots of text data by their smartphones, these data are building blocks for now-days big language models. However, it is shown that most language models suffer from ethic problems, since they may leak users' personal information in an unexpected way. 

Recent efforts in federated learning have been devoted to interdisciplinary areas: not only machine learning is required, but also techniques from distributed optimization, statistics, cybersecurity, communication, systems, cryptography and many more. Meanwhile, the data ranges from structured to unstructured format, which is not limited to tabulars, time series and images.  Among most federated learning studies, Google has led the use of federated learning in Natural Language Processing through Gboard mobile keyboard, Pixel phones and Android Messages. While Google has launched several applications on langauge modeling tasks, Apple is using FL for wake-up word detection in Siri, doc.ai is developing cross-device FL solutions for biomedical research, and Snips has introduced cross-device FL for hotword detection.

In this paper, we take a survey on the existing FL algorithms for Natural Language Processing (NLP). Starting from language modeling, we will review current federated learning algorithms on various NLP tasks: classification, recommendation, speech recognition and health text mining and others . We organize the survey as follows: in Section 2, basic federated learning concepts, frameworks, optimization toward non-IID data, privacy are discussed. Section 3 reviews federated learning in NLP.  Section 4 discusses the common evaluation aspects and tools. Section 5 highlights current research challenges and some future directions. Section 6 gives the conclusion.
\section{Federated learning}
In this section, we first review basics of federated learning, including the problem setup, non-iid data distribution, frameworks, optimization algorithms and privacy preservation. Then, we extend federated learning to other distributed machine learning paradigms and discuss their difference. 
\paragraph{Problem formulation} In this wrok, we consider the following distributed optimization process:
\begin{center}
    $\min_\textbf{w}\{\mathcal{L}(\textbf{w}) = \sum_{k=1}^N p_k \mathcal{L}_k(\textbf{w})\}$,
\end{center}
where $N$ is the total number of user devices, $p_k$ is the weight of the $k$-th device such that $p_k \geq 0$ and $\sum_{k=1}^{N}p_{k}=1$. Suppose the $k$-th device has the amount of $n_k$ training data: $\textbf{x}_k = (x_{k,1},x_{k,2}, ..., x_{k,n_{k}})$. The local training objective $\mathcal{L}_k(.)$is defined by:
\begin{center}
    $\mathcal{L}_{k}(\textbf{w}) = \frac{1}{n_k}\sum_{j=1}^{n_k} l(\textbf{w}; x_{k,j})$,
\end{center}
where $l(.;.)$ is a user-specified loss function.

\paragraph{Non-IID data and Learning Strategies} 
Typical centralized supervised learning algorithms have the IID assumption, i.e., the training and test data is independently identically distributed. In decentralized settings like federated learning, non-IID poses a challenge because the different data distribution result in significant skewness across devices or locations. Non-IID data among devices/locations encompass many different forms. There can be skewed distribution of features (probability $\mathcal{P}(x)$), labels (probability $\mathcal{P}(y)$), or the relationship between features and labels (e.g., varying $\mathcal{P}(y|x)$ or $\mathcal{P}(x|y)$) among devices/locations. Previous reviews categorized this this as horizontal, vertical and hybrid data partitions in Federated Learning. In this review, we focus on skewed distribution of labels, i.e.,  $\mathcal{P}_{P_i}(y) \not= \mathcal{P}_{P_j}(y)$ for different data partitions $P_i$ and $P_j$. 

Previous study has shown DNN models with batch normalization suffer from Non-IID data \cite{hsieh2020non}, the accuracy of FL reduces significantly by up to 55\% for neural networks trained for highly skewed non-IID data \cite{zhao2018federated}, where each clinet device trains only on a single class of data. 

Common techiniques to deal with non-IID include:
\begin{itemize}
    \item data augmentation: create a common dataset which can be shared globally, the dataset can come from a publicly available proxy data source \cite{zhao2018federated}, or perhaps a distillation of the raw data following \cite{wang2018dataset}.
    \item schedule client participation during training : FedFMC, FedCD, cluster similar devices / multi-center/ hirarchical clustering of local updates, FedPD, adapts the communication frequency of decentralized learning algorithms to the (skew-induced) accuracy loss between data partitions \cite{hsieh2020non} 
    
    \item greater number of models, but more communication cost:
    
    \item ensemble: similar to  scheduling 
    
    \item regularization on the server, e.g. FedAwS, server imposes a geo- metric regularizer after each round to encourage classes to be spreadout in the embedding space
    
    \item personalized FL/ continual local training, based on MAML
    
\end{itemize}

\paragraph{Optimization}
While a variety of studies have made assumptions for the per-client optimization functions in the IID setting, we review basic convergence results for $H$-smooth convex functions under the assumption that the variance of the stochastic gradients is bounded by $\sigma ^2$. Given the following notations in a standard FL setting: $N$ is the total number of clients, $M$ is the number of participated clients per round, $T$ is the total number of communication rounds, $K$ is the local SGD steps per round. Federated averaging can conducted in either of the following two settings: one is to keep $x$ fixed in local updates during each round and compute a total of $KM$ gradients at the current $x$, in order to run accelerated minibatch SGD, the convergence rate is then upper bounded by $O(\frac{H}{T^2}+\frac{\mu}{\sqrt{TKM}})$. The other is to ignore all but 1 of the M active clients, which allows sequential SGD to run for $KT$ steps, this approach has an upper bound of $O(\frac{H}{(TK)^2}+\frac{\mu}{\sqrt{TK}})$. As in the non-IID settings, key assumptions are given for inter-client gradient, local functions on each client and other participation constraints. A detailed discussion of different convergence rates for non-IID setting can be found in \cite{kairouz2019advances}.
\paragraph{Frameworks}
There are three basic frameworks for FL: centralized, decentralized and heterogeneous. In the centralized federated learning setting, a central server is used to orchestrate the different steps of the algorithms and coordinate all the participating nodes during the learning process. The server is responsible for the nodes selection at the beginning of the training process and for the aggregation of the received model updates. Since all the selected nodes have to send updates to a single entity, the server may become a bottleneck of the system. Most NLP applications like keyboard word prediction is using the cetralized setting. 
In the decentralized federated learning setting, the nodes are able to coordinate themselves to obtain the global model. This setup prevents single point failures as the model updates are exchanged only between interconnected nodes without the orchestration of the central server. Nevertheless, the specific network topology may affect the performances of the learning process.  Most blockchain-based federated learning falls into the decentralized setting. An increasing number of application domains involve a large set of heterogeneous clients, e.g., mobile phones and IoT devices. Most of the existing Federated learning strategies assume that local models share the same global model architecture. Recently, a new federated learning framework named HeteroFL was developed to address heterogeneous clients equipped with very different computation and communication capabilities. The HeteroFL technique can enable the training of heterogeneous local models with dynamically-varying computation complexities while still producing a single global inference model.
\paragraph{Privacy}
In most FL settings, privacy preservation is conducted to make sure users' information is not leaked during the learning process. Physically, local data is not allowed to leave end users' devises. However, it is still possible to reconstruct the original data by taking the model weights or gradients. Therefore, we consider privacy preservstion on three aspects in FL: users' personal information, local data and machine learning models. Take the smart phone keyboard next word prediction as an example, users personal information refers to facts about their location, name, sex as well as hidden information like keyboard typing pattern. Local data then concludes the messages, photos and videos in their phones, while a machine learning model could be the a language model which predicts the next word given some preceding words. Users' personal information is often correlated with the local data, e.g., a person's age can be inferred with his/her chat messages. 

Common techniques for privacy preservation of user data includes: differential privacy \cite{dwork2008differential}, secure Multi-Party Computation \cite{goldreich1998secure}, homomorphic encryption \cite{naehrig2011can} and trusted execution environments \cite{ekberg2013trusted}. Verifiability enables parties to prove that they have executed their parts of a computation faithfully. Techniques for verifiability include both zero knowledge proofs (ZKPs) \cite{feige1988zero} and trusted execution
environments (TEEs) \cite{ekberg2013trusted}. As for model attack, adversarial learning techniques can be leveraged in FL setting, 

\section{Federated learning in NLP}
\subsection{Language modeling}
A language model (LM) refers to a model that provides probabilities of word sequences through an unsupervised distribution estimation. As an essential component for NLP systems,  LM is utilised in a variety of NLP tasks, i.e., machine translation, text classification, relation extraction, question-answering, etc. In FL, most LMs are deployed on a virtual mobile keyboard, i.e., the Google Keyboard (Gboard). Thereby, recent literature are mostly produced by authors from Google, LLC. Recent works on language modelling in Federated NLP mainly target on solving a word-level LM problem in mobile industry. That is mobile keyboard suggestion, which is a well representative of federated NLP applications. To improve mobile keyboard suggestions, federated NLP models aim to be more reliable and resilient. Existing models offer quality improvements in typing or even expression (e.g., emoji) domains, such as next-world predictions, emoji predictions, query suggestions, etc.

Considering the characteristics of mobile devices, a decentralized computation approach is constrained by computation resource and low-latency requirement. A mobile device has limited RAM and CPU budgets, while we expect keyboards to provide a quick and visible response of an input event within 20 milliseconds. Thereby, the model deployed in client sides should perform fast inference. 

Most works \cite{Yang2018APPLIEDFL,Hard2018FederatedLF,Ramaswamy2019FederatedLF,Chen2019FederatedLO,Ji2019LearningPN,Stremmel2020PretrainingFT,Thakkar2020UnderstandingUM} consider variants of LSTMs \cite{Hochreiter1997Long} as the client model. Given the limited computation budget on each device, we expect the parameter space of a neural language model to be as small as possible without degrading model performance. CIFG \cite{Greff2017LSTMAS} is a promising candidate to diminish LM’s complexity and inference-time latency. It employs a single gate to harness both the input and recurrent cell self-connections. In such a way, the amount of parameters is downsized by 25 $\%$ \cite{Greff2017LSTMAS}. \cite{Hard2018FederatedLF} leverages CIFG for next word predictions, and simplifies the CIFG model by removing peephole connections. To further optimise the model in size and training time, they ties input embedding and CIFG output projection matrices. \cite{Ramaswamy2019FederatedLF} applies a pretrained CIFG network as an emoji prediction model. In particular, the pretraining process involves all layers, excluding the output projection layer, using a language learning task. To enhance performance, the authors enable embedding sharing between inputs and outputs. The pretraining of LM exhibits fast convergence for the emoji model. \cite{Chen2019FederatedLO} employs a character-level RNN \cite{1998A}, targeting on out-of-vocabulary(OOV) learning tasks, under FL settings. Specifically, they use CIFG with peephole connections and a projection layer. The projection layer diminishes the dimension of output and accelerates the training. They use multi-layer LSTMs to enhance the representation power of the model, which learns the probability of word occurrence. GRU \cite{2014Learning} is another simpler variant of the LSTM. \cite{Ji2019LearningPN} leverage GRU as the neural language model for mobile keyboard next-word predictions. Similar to CIFG, it reduces the model complexity on parameter spaces without hurting the model performance. To downsize the amount of trainable parameters, they also apply tied embedding in the embedding layer and output layer by share of the weights and biases.

\cite{Yang2018APPLIEDFL} proposes another LM for keyboard query suggestions to reduce the burden of training LSTMs. Specifically, they train a LSTM model on the server for generating suggestion candidates, while merely federated training a triggering model, that decides the occurrence of the candidates. The triggering model uses logistic regression to infer the probability of a user click, significantly lessening the computation budgets in comparison of RNN models. \cite{chen-etal-2019-federated} also states the direct use of RNN is not the proper means to decode due to its large parameters size, which further causes slow inference. Hereby, they propose to leverage a n-gram LM that derived from a federated RNN for decoding. In particular, they overcome large memory footprints problem and enhance model performance by introducing an approximation algorithm based on SampleApprox \cite{Suresh2019ApproximatingPM}. It approximates RNN models into n-gram models. Still, they use CIFG and group-LSTM (GLSTM) \cite{2017Factorization} for approximation. While, GPT2 \cite{0Language} is one of the state-of-the-art transformer-based LMs with 1.5 billion parameters. Considering its performance on centralized language modeling tasks, \cite{Stremmel2020PretrainingFT} uses GPT2 as LM. They propose a dimensionality reduction algorithm to downsize the dimension of GPT2 word embedding to desired values (100 and 300).

For federated optimization, existing federated optimization algorithms differ in client model aggregation on the server-side. In federated language modeling, most existing works \cite{Yang2018APPLIEDFL,Hard2018FederatedLF,chen-etal-2019-federated,Ramaswamy2019FederatedLF,Chen2019FederatedLO,Stremmel2020PretrainingFT,Thakkar2020UnderstandingUM} use FedAvg as the federated optimization algorithm. Another optimization strategy, called FedAtt, has also shown its feasibility and validity in language models \cite{Ji2019LearningPN}.

In FedAvg, gradients, that computed locally over a large population of clients, are aggregated by the server to build a novel global model. Every client is trained by locally stored data and computes the average gradient with the current global model via one or more steps of SGD. Then, it communicates model updates with the server. The server performs the weighted aggregation of the client updates to build a novel global model. Client updates are immediately abandoned on the sever once the accumulation is completed. \cite{Hard2018FederatedLF} trains the global model from scratch in the server, using FedAvg. Specifically, the initial global model has either been randomly initialized or pretrained on proxy data. However, it increases the federated training rounds on clients. Thereby, it leads to a high communication and computation costs in FL. They also use SGD as the server-sided optimizer for training. They found Adam and AdaGrad provide no beneficial improvement on convergence. \cite{Stremmel2020PretrainingFT} introduces a novel federated training approach, called central pre-training with federated fine-tuning. To address the drawback in \cite{Hard2018FederatedLF}, the server pretrains a model with centralized and public data as the global model at the initial time. Each clients then obtains the pretrained weights as the initial weights, and later trained on local data in a federated fashion. But the improvement is limited to large network, i.e., GPT2. They also propose a pretrained word embedding layer for federated training, which only enhance accuracy for the large word embedding network (i.e., GPT2). Whereas, with the combination of pretraining models, it harms the performance. They leverage Adam as the optimizer for training. \cite{Chen2019FederatedLO} uses momentum and adaptive L2-norm clipping on each client’s gradient in FedAvg, leading to a faster convergence. The authors argue momentum and adaptive clipping performed on gradients improves the robustness of model convergence and performance. \cite{Thakkar2020UnderstandingUM} also uses clipping for regularization in FedAvg by setting the upper bound of user updating to constrain each client contribution (i.e., clipping). In addition, \cite{Ramaswamy2019FederatedLF} founds using momentum with Nesterov accelerated gradients significantly outperforms using SGD as server optimizer, in terms of convergence rate and model performance. \cite{chen-etal-2019-federated} applies Nesterov momentum as both the local and the server optimizer.

\cite{Ji2019LearningPN} first introduces the attention mechanism into federated aggregation of client models. This optimization algorithm is referred as Attentive Federated Aggregation (FedAtt). It is a layer-wise soft attention mechanism applied on the trained parameters of the NN model. Intuitively, the federated optimization algorithm learns to optimize the global model by providing a good generalization on each client model for a quick local adaptation. Hereby, it reduces local training rounds and saves the computation budgets, further accelerating the learning process. The generalization in FedAtt is decided by the similarity between each client and the server, and the relative importance of each client. For a good generalization, they minimise the weighted summed distance of each client model and the global model on parameters spaces. They introduce attentive weights as the weights of the client models. Particularly, the attentive weight of each client model is a non-parametric attention score derived from each layer of NN.  Differ from pre-trained FedAvg, FedAtt finds a well-generalized global model on each federated training round by iteratively updating parameters. Consequently, it further lessens the federated communication budgets. For local training, the client-sided optimizer is momentum. While, for global parameters updates, they uses SGD.

The existing works on federated language modeling mainly contribute on optimizing model aggregation process, but not focusing on privacy preserving approach. Adding privacy preserving techniques into federated optimization process is seen as a bonus, rather than an essential means of privacy guarantees. In Federated LMs, the commonly used privacy preserving technique is differential privacy (DP) \cite{2006Calibrating}. A DP algorithm is expected to characterize the underlying probability distribution without compromising personally identifiable data. In general, it injects calibrated noise into the aggregated data while not affecting the outcomes. Most DP approaches are used for user-level privacy guarantees. In FL, we define user-level DP as a privacy guarantees, to preserve the trained models with or without the presence of any one client's data. DP usually serves on the client sides before model aggregation \cite{Wei2020FederatedLW}. \cite{Ji2019LearningPN} integrates a randomized mechanism in FedAtt optimization by introducing a white noise with the mean of 0 and the standard deviation $\sigma$. They also introduce a magnitude coefficient $\beta \in (0,1]$ to govern the effect of the randomization in FL. However, the level of its DP guarantees is unrevealed. Hereby, it fails to show the trade-off between data utility and privacy protection for its privacy-preserving countermeasure implementation. \cite{Thakkar2020UnderstandingUM} incorporates the Gaussian mechanism in FedAvg to cope with the user-based heterogeneity of data in language models. In particular, it perform DP guarantees by adding Gaussian noise with a noise multiplier of 1, after clipping. They argue a high level of DP guarantees exhibits a notable reduction in unintended memorization, caused by heterogeneity of training data.

\subsection{Classification}

Text Classification is procedure of identifying the pre-defined Text Classification is the procedure of identifying the pre-defined category for varied-length of text \cite{aggarwal2012survey}. It can be extended to many NLP applications including sentiment analysis, question answering and topic labeling .
Traditional text classification tasks can be deconstructed into four steps: text preprocessing, dimension reduction, classification and evaluation. 
Though the deep learning models have achieved state-of-the-art results in text classification \cite{minaee2021deep}, uploading or sharing text data to improve model performance is not always feasible due to different privacy requirements of clients. For example, financial institutions that wish to train a chatbot for their clients cannot be allowed to upload all text data from the client-side to their central server due to strict privacy protection statements. Then applying the federated learning paradigm is an approach to solve the dilemma due to its advances in privacy preservation and collaborative training.  In which, the central server can train a powerful model collaboratively with different local labeled data at client devices without uploading the raw data considering increasing privacy concerns in public.

However, there are several challenges for applying federated learning to text classification tasks in NLP. One is to design proper aggregating algorithms to handle the gradients or weights uploaded by different client models. Traditional federated learning can be considered as a special paradigm of distributed learning, thus aggregating algorithms, such as FedAvg \cite{DBLP:journals/corr/McMahanMRA16}, FedAtt \cite{Ji2019LearningPN} has been proposed to generalize the model on the central server. Considering the unevenly distributed data at different client devices and different amounts of data at the different local datasets. 
\cite{zhu2020empirical} has attempted the text classification using the standard FedAvg algorithm to update the model parameter with local trained models. It uses different local datasets to pre-train the word embeddings, and then concatenate all word embeddings. After filtering the widths and feature maps from the concatenated word embeddings, the max-over-time pooling was used to aggregate the features, thus getting vectors with the same length. Finally, they use softmax activation on the fully connected layer, it will translate the vectors to the final sentence classification results (categories).
Later, scientists from the Machine learning area brought in new approaches of uploading and aggregating, for example,  using Knowledge distillation \cite{hinton2015distilling}. \cite{hilmkil2021scaling} however use fine-tuning instead of FedAvg to update parameters.  \cite{lin2021ensemble} average the logits outputs from the last layer of the model instead of directly take the average of model parameters. It then uses knowledge distillation to learn the knowledge from the client devices instead of traditional.

In addition, model compression has been introduced to federated text classification tasks due to the dilemma of computation restraints on the client-side. They attempted to reduce the model size on the client-side to enable the real application of federated learning. The computation restriction on the client devices limits the application of traditional FL. For example, 4-layer BERT or 6-layer BERT is still too large for mobile devices such as smartphones. The scholars then focus to perform the model compression while still following the federated learning paradigm. The knowledge distillation then has been applied to transfer local model information while keeping the model size small at the local devices in \cite{sattler2020communication}. It utilises knowledge distillation instead of model parameter fusion to update parameters. The soft-label predictions on a public distillation dataset are sent to the central model to be distilled. Thus, the central model can learn the local knowledge on client devices through distilling the logits of different client models without sharing or uploading the local model parameters and gradients.

To ensure the privacy preservation of FL while keeping the communication, the encryption of data is one of the top priority considerations in applying federated learning in NLP. Encryption on communication between edge-device and central server is a standard approach in federated learning to preserve privacy for end-users on edge devices. \cite{zhu2020empirical} adds encryption on client-central server communication using differential privacy. It used the approach \cite{wan2021robust} proposed the attack-adaptive aggregation which prevent the attack at the central server aggregation module.

To overcome the communication dilemma of FL, one-shot or few-shot federated learning was proposed to allow the central server can successfully train the central model with only one or a few rounds of communication under poor communication scenarios. However, the shared data restriction of federated learning is still left to be solved. Considering the trend of higher restriction of data sharing and uploading, it will be harder to get a sufficient size of data shared to both central servers and client servers. In this way, the knowledge distillation cannot be used to solve the model compression problem in federated learning.
\cite{zhou2020distilled} reduced the communication of previous federated learning by utilising the soft labels dataset distillation mentioned in  \cite{guha2019one} and \cite{sucholutsky2019soft}. It thus successfully extend the soft-labeling methods to two new techniques: soft-reset and random masking, and then successfully using the dataset distillation \cite{wang2018dataset} to realise the one-round communication federated learning for text classification tasks. Each client in  \cite{zhou2020distilled} distils their local dataset to a much smaller synthetic one, and then only uploads the small-sized synthetic dataset to the server. Thus, no gradients or weights is transmitting from the client model to the central server model. The distilled dataset can be as small as one data sample per category, in this way the communication in federated learning can be reduced to as low as one round. 

\subsection{Speech Recognition}
Speech recognition is the task of recognising speech within audio and converting it into text. Voice assistants such as Amazon Alexa or Apple Siri use on-device processing to detect wake-up words (e.g. "Hey Siri"), which is a typical usage for speech recognition on smartphones. Only when the wake-up words are detected, further processing like information retrieval or question answering will be running on the cloud. Methods for speech recognition include dynamic time wraping \cite{muller2007dynamic}, Hidden Markov Models \cite{rabiner1986introduction} and modern end-to-end deep neural models \cite{graves2014towards}. More recently, wav2vec \cite{baevski2020wav2vec} masks the speech input in the latent space and solves a contrastive task defined over a quantization of the latent representations which are jointly learned, this method demonstrates the feasibility of speech recognition with limited amounts of labeled data.

On device wake-up word detectors face two main challenges: First, it should run with minimal memory footprint and computational cost. Second, the wake word detector should behave consistently in any usage setting, and show robustness to background noise. \cite{zhang2017hello} performed neural network architecture evaluation and exploration for running keyword spotting on resource-constrained microcontrollers, they showed that it is possible to optimize these neural network architectures to fit within the memory and compute constraints of microcontrollers without sacrificing accuracy. \cite{leroy2019federated} investigated the use of federated learning on
crowdsourced speech data to learn a resource-constrained wake word detector. They showed that a revisited Federated Averaging algorithm with per-coordinate averaging based on Adam in place of standard global averaging allows the training to reach a target stopping criterion of 95\% recall per 5 FAH within 100 communication rounds on their crowdsourced dataset for an associated upstream communication costs per client of 8MB.
They also open sourced the Hey Snips wake word dataset \footnote{http:// research.snips.ai/datasets/keyword-spotting}. \cite{yang2020decentralizing} proposed a decentralized feature extraction approach in federated learning to address privacy-preservation issues for speech
recognition, which is built upon a quantum convolutional neural network (QCNN) composed of a quantum circuit encoder for feature extraction, and a recurrent neural network (RNN) based end-to-end acoustic model (AM). The proposed decentralized framework takes advantage of the quantum learning progress to secure models and to avoid privacy leakage attacks. \cite{guliani2020training} introduced a framework for speech recognition by which the degree of non-IID-ness can be varied, consequently illustrating a trade-off between model quality and the computational cost of federated training. They also showed that hyperparameter optimization and appropriate use of variational noise are sufficient to compensate for the quality impact of non-IID distributions, while decreasing the cost.

\subsection{Sequence Tagging}
Sequence tagging, e.g. POS tagging, Named Entity Recognition, plays an important role in both natural language understanding and information extraction. Statistical models like Hidden Markov Model and Conditional Random Fields were heavily used, modern approaches rely on deep representations from Recurrent Neural Net, Convolution Neural Net or transformer like architectures. A few recent works focus on biomedical Named Entity Recognition in the federated setting. \cite{liu2020federated} pretrained and fine tuned BERT models for NER tasks in a federated manner using clinical texts, their results suggested that conducting pretraining and fine tuning in a federated manner using
data from different silos resulted in reduced performance compared with training on centralized
data. This loss of performance  is mainly due
to separation of data as ”federated communication
loss” .  Given the limit of data access, the experiments were conducted with clinical notes from a single healthcare system to simulate different silos. \cite{ge2020fedner} proposed a FedNER method for medical NER, they decomposed the medical NER model in each platform into
a shared module and a private module. The private
module was updated in each platform using the local
data to model the platform-specific characteristics.
The shared module was used to capture the shareable
knowledge among different platforms, and was updated in a server based on the aggregated gradients from multiple platforms.  The private module consists of two top layers in our medical NER model, i.e, Bi-LSTM and CRF, which aim to learn
platform-specific context representations and label
decoding strategies.  The private module
was only trained with local data and exchange neither its parameters nor gradients. The shared module consisted
of the other bottom layers in our NER model, such
as the word-level CNN and all types of embedding.
Different from the private module, the shared one
mainly aims to capture the semantic information in
texts. \cite{sui2020feded} introduced a privacy preserving medical relation extraction model
based on federated learning, they leveraged a strategy based on knowledge distillation. Such a strategy uses the uploaded predictions of ensemble local models to train the central model without requiring uploading local parameters.

\subsection{Recommendation}
Recommendation systems are heavily data-driven. Typical recommendation models use collaborative filtering methods \cite{suganeshwari2016survey}, in which past user item interactions are sufficient to detect similar users and/or similar items and make predictions based on these estimated proximities. Collaborative filtering algorithms can be further divided into two sub-categories that are generally called memory based and model based approaches. Memory based approaches directly works with values of recorded interactions, assuming no model, and are essentially based on nearest neighbours search (for example, find the closest users from a user of interest and suggest the most popular items among these neighbours). Model based approaches assume an underlying “generative” model that explains the user-item interactions and try to discover it in order to make new predictions. Unlike collaborative methods that only rely on the user-item interactions, content based approaches \cite{pazzani2007content} use additional information about users and/or items. If we consider the example of a movies recommendation system, this additional information can be, for example, the age, the sex, the job or any other personal information for users as well as the category, the main actors, the duration or other characteristics for the movies (items).

Given different partitions of users and items, federated recommendation models can be horizontal, vertical or transfered. In horizontal federated recommendation systems, items are shared but users belong to different parties. A typical work is Federated Collaborative Filter (FCF) \cite{ammad2019federated} proposed to use a central server to keep the item latent factor matrix, while the user latent factors are stored locally on each device. In the training time, the server distributes the item latent factor to each party, the participants update their user latent factor by local rating matrix data and send the item latent factor updates back to the server for aggregation. To avoid the inter trust problem, \cite{hegedHus2019decentralized} introduced a fully decentralized setting where participants have full access to the item latent factor and communicate with each other to update the model. Moreover, meta learning has been used for personalized federated recommendation. \cite{fallah2020personalized} designed a meta learner to learn generalized model parameters for each participant, then each participant's recommendation is regarded as a personalized task where a support set is used to generate the recommendation model and the gradient is computed on a query set. \cite{jalalirad2019simple} introduced another fedrated meta learning algorithm for recommendation, in which separate support and query sets are not necessary. Their approach performs relatively well within less amount of training episodes. Besides, \cite{li2016difacto} proposed DiFacto, which is a distributed factorization method and addressed the efficiency problem when it comes to large scale user item matrices. In comparison, vertical federated systems have been designed for feature distributed learning problem where participants hold different feature sets. \cite{hu2019fdml} proposed an
asynchronous stochastic gradient descent algorithm. Each party could use an
arbitrary model to map its local features to a local prediction. Then local predictions from different parties are aggregated into a final output using linear and
nonlinear transformations. The training procedure of each party is allowed to be
at various iterations up to a bounded delay. This approach does not share any
raw data and local models. Therefore, it has fewer privacy risks. Besides, for a
higher level of privacy, it can easily incorporate the DP technique. Similar to
horizontal FedRec, there are also works that further utilize cryptography techniques.
\cite{cheng2019secureboost} presented a secure gradient-tree boosting algorithm. This algorithm
adopts HE methods to provide lossless performance as well as preserving privacy.
And \cite{slavkovic2007secure} proposed a secure linear regression algorithm. MPC protocols are
designed using garbled circuits to obtain a highly scalable solution.
Parties of vertical FedRec could also be two recommenders with different item
sets. For instance, a movie RecSys and a book RecSys have a large user overlapping
but different items to recommend. It is assumed that users share a similar
taste in movies with books. With FedRec, the two parties want to train better
recommendation algorithms together in a secure and privacy-preserving way.
\cite{shmueli2017secure} proposed a secure, distributed item-based CF method. It jointly improves
the effect of several RecSys, which offer different subsets of items to the same
underlying population of users. Both the predicted ratings of items and their
predicted rankings could be computed without compromising privacy nor predictions’ accuracy. We refer readers to \cite{yang2020federated} for more detailed discussion for federated recommendation systems.

\subsection{Health Text Mining}
Federated learning has emerged as an important framework for health text mining, due to the privacy concern among different hospitals and medical organizations. Besides, most health data exhibits systemic bias towards some specific groups or patterns, e.g. hospitals, diseases and communities. Again, this non-IID issue raises big challenges when applying federated learning into heath text mining tasks. There have been some tasks that were studied in federated learning setting in healthcare, including patient similarity learning \cite{lee2018privacy}, patient representation learning and phenotyping \cite{kim2017federated,liu2019two}, predictive or classification modeling \cite{brisimi2018federated,huang2019patient,sharma2019preserving}, biomedical named entity recognition. 

Specifically, \cite{liu2019two} designed a two-stage federated approach for medical record classification. In the first stage, they pre-trained a patient representation model by training an neural network to predict ICD and CPT codes from the text of the notes.  
In the second stage, a phenotyping machine learning model was trained in a federated manner using clinical notes that are distributed across multiple sites for the target phenotype. In this stage, the notes mapped to fixed-length representations from stage one are used as input features and whether the patient has a certain disease is used as a label with one of the three classes: presence, absence or questionable. \cite{lincy2020early} proposed a simple federated architecture for early detection of Type-2 diabetes. After comparing the proposed
federated learning model against the centralised approach, they showed that the federated learning model ensures significant privacy over
centralised learning model whereas compromising accuracy for a subtle extend.  To cope with the imbalanced and non-IID distribution inherent in user's monitoring data, \cite{wu2020fedhome} designed a generative convolutional autoencoder (GCAE), which aims to achieve accurate and personalized health monitoring by refining the model with a generated class-balanced dataset from user's personal data. It is noticed that GCAE is lightweight to transfer between the cloud and edges, which is useful to reduce the communication cost of federated learning in FedHome. \cite{stripelis2021scaling} described a federated approach on a brain age prediction model on structural MRI scans distributed across multiple sites with diverse amounts of data and subject (age) distributions. In these heterogeneous environments, a Semi-Synchronous protocol provides faster convergence. 

\subsection{Other NLP Tasks}
More recently, FedNLP provided a research-oriented benchmarking framework for advancing federated learning (FL) in natural language processing (NLP). It uses FedML repository as the git submodule. In other words, FedNLP only focuses on adavanced models and dataset, while FedML supports various federated optimizers (e.g., FedAvg) and platforms (Distributed Computing, IoT/Mobile, Standalone). A text generation example can also be found in TensorFlow Tutorial \footnote{https://colab.research.google.com/github/tensorflow/federated/blob/master/docs/tutorials \\ /federated\_learning\_for\_text\_generation.ipynb\#scrollTo=iPFgLeZIsZ3Q}. So far, we have not found any work on other generation works on Machine Translation and Summatization.
\section{Benchmarks}
\subsection{Evaluation Aspects}
\paragraph{Model Evaluation}In principle, FL based NLP models should not only be evaluated against traditional performance metrics (such
as model accuracy), but also the change of model performance with different system and data settings. Various systems settings consider the number of nodes, the weight of the nodes, the quality of the nodes. While the data setting focus on different data distribution caused by either label bias or feature bias. 

\paragraph{Communication Evaluation} There is no doubt that the communication rounds of nodes play an important role
in the performance of the model. Due to the uncertainty of the federated network, communication is huge resource consumption. There is always a natural trade off between computation and communication among the nodes and server. 

\paragraph{Privacy Evaluation}
The goal of privacy metrics is to measure the degree of privacy enjoyed by users in a system and the amount of protection offered by privacy-enhancing technologies. \cite{wagner2018technical} discussed a selection of over eighty
privacy metrics and introduce categorizations based on the aspect of privacy they measure, their required
inputs, and the type of data that needs protection. In general, privacy metrics can be classified with four common characteristics: adversary models, data sources, inputs and output meatures.

\paragraph{User Response Evaluation} Apart from the above automatic evaluation methods, on-line FL-based NLP models also consider the response from users, e.g. FL language models take next word click rate as an important metric, FL recommendation systems would not only like to keep old customers but also attract new customers. 

\subsection{Tools}
There are a few tools for common federated learning, including PySyft \footnote{https://github.com/OpenMined/PySyft}, TFF \footnote{https://www.tensorflow.org/federated}, FATE \footnote{https://github.com/FederatedAI/FATE}, Tensor/IO \footnote{https://doc-ai.github.io/tensorio/}, FedML \footnote{https://github.com/FedML-AI/FedML} and FedNLP \footnote{https://github.com/FedML-AI/FedNLP}.
PySyft  decouples private data from model training using federated learning, DP and MPC within PyTorch. With TFF, TensorFlow provides users
with a flexible and open framework
through which they can simulate distributed
computing locally. FATE support the Federated AI
ecosystem, where a secure computing
protocol is implemented based on
homomorphic encryption and MPC. Tensor/IO is a lightweight crossplatform
library for on-device
machine learning, bringing the power
of TensorFlow and TensorFlow Lite
to iOS, Android, and React native
applications.
\section{Challenges and Future Directions}
\subsection{Algorithm-Level}
\paragraph{Big language models} Since the paradigm pre-training + fine tuning has dominated most NLP tasks, pre-trained language models such as BERT and GPT are useful and transferable to develop downstream tasks. Often times the larger the pre-trained language model is, the more likely it will be for downstream model performance. However, in the FL setting, it is not possible to allocate large size language models like GPT-3 on the participants. Technique like knowledge distillation could be useful, it remains unknown whether downsized language can maintain the performance.
\paragraph{Non-iid Data distributions} Real world data from different participants is always non-iid, the challenge is how to learn from high quality data and labels.  Given a fixed annotation budget, active learning may be leveraged to not only select significant data points, but also actively choose worthwhile participants. Furthermore, weakly supervised learning and meta learning algorithms may also be used to use more unlabeled data from different participants. 
\paragraph{Personalization} Personalized FL can be
viewed as an intermediate paradigm between the server-based
FL paradigm that produces a global model and the local
model training paradigm. The challenge is to strike a careful balance between local task-specific knowledge and shared knowledge useful for the generalization properties of FL models. For most deep NLP models, techniques like early shaping, sample weighing and transfer learning can be explored.

\subsection{System-Level}
\paragraph{Spatial Adaptability}  This refers to the ability of the FL system to handle variations across client data sets as a result of
(i) the addition of new clients, and/or (ii) dropouts and stragglers. These are practical issues prevalent in complex edge computing environments, where there is significant variability in hardware capabilities in terms of computation, memory, power and network connectivity

\paragraph{Computation Communication Trade-off} Frequent and large scale deployment of updates, monitoring, and debugging for FL NLP models is challenging. The trade-off between local and global update frequency, as well as the communication frequency can be explored.
\paragraph{Privacy concern} Even though FL assumes the data never leave the device, it is still possible to reconstruct the original data by taking the model weights or gradients. Privacy preservation on three aspects in FL can be explored: users' personal information, local data and machine learning models.
\section{Conclusion}
In this paper, we review common NLP tasks in the FL setting, including language modeling, text classification, speech recognition, sequence tagging, recommendation, health text mining and other tasks. In general, the performance federated NLP models still lie behind that of centralized ones. Also, large scale pre-trained language models and advanced privacy preservation techniques have not widely been used in the FL based NLP, which could be the potentials for future research. We point out both algorithm and system level challenges for FL based NLP models. In the future, we will further evaluate representative NLP models (e.g. transformers) in the FL environment and give more comparable insights on real world applications. 
%
%
%
\bibliographystyle{splncs04}
\bibliography{bib.bib}

\end{document}